\documentclass[runningheads]{llncs}
\usepackage[T1]{fontenc}
\usepackage{graphicx}
\usepackage{algorithm} 
\usepackage{algorithmicx}
\usepackage{algpseudocode} 
\usepackage{amssymb}
\usepackage{booktabs}
\usepackage{multirow}

\begin{document}
\title{SimMining-3D: Altitude-Aware 3D Object Detection in Complex Mining Environments:\\ A Novel Dataset and ROS-Based Automatic Annotation Pipeline\thanks{Supported by Rio Tinto Centre for Mine Automation, Australian Centre for Robotics.}}
\titlerunning{SimMining-3D}
%
\author{Mehala Balamurali\inst{1}\orcidID{0000-0003-0083-5772} \and
Ehsan Mihankhah \inst{2}}
%

%
\institute{Australian Centre for Robotics, University of Sydney, Australia\\
\email{mehala.balamurali, ehsan.mihankhah@sydney.edu.au}}
\maketitle              
\begin{abstract}
Accurate and efficient object detection is crucial for safe and efficient operation of earth-moving equipment in mining. Traditional 2D image-based methods face limitations in dynamic and complex mine environments. To overcome these challenges, 3D object detection using point cloud data has emerged as a comprehensive approach. However, training models for mining scenarios is challenging due to sensor height variations, viewpoint changes, and the need for diverse annotated datasets.

This paper presents novel contributions to address these challenges. We introduce a synthetic dataset SimMining-3D\cite{ref_article10} specifically designed for 3D object detection in mining environments. The dataset captures objects and sensors positioned at various heights within mine benches, accurately reflecting authentic mining scenarios. An automatic annotation pipeline through ROS interface reduces manual labor and accelerates dataset creation.

We propose evaluation metrics accounting for sensor-to-object height variations and point cloud density, enabling accurate model assessment in mining scenarios. Real data tests validate our model's effectiveness in object prediction. Our ablation study emphasizes the importance of altitude and height variation augmentations in improving accuracy and reliability.

The publicly accessible synthetic dataset\cite{ref_article10} serves as a benchmark for supervised learning and advances object detection techniques in mining with complimentary pointwise annotations for each scene. In conclusion, our work bridges the gap between synthetic and real data, addressing the domain shift challenge in 3D object detection for mining. We envision robust object detection systems enhancing safety and efficiency in mining and related domains.

\end{abstract}
\section{Introduction}
Effective object detection is vital for ensuring the safe and efficient operation of earth-moving equipment in the mining industry. However, traditional 2D image-based methods often encounter limitations in dynamic and complex mine environments where objects can be occluded or obscured. To overcome these challenges, 3D object detection utilizing point cloud data provides a more comprehensive representation of objects and the environment, leading to improved accuracy and efficiency.

Training models for 3D object detection poses unique challenges, including adapting pretrained models to new datasets and effectively handling viewpoint variations. Additionally, collecting real-world data for training in mining scenarios can be particularly challenging due to complex terrains, cluttered surroundings, safety concerns, and logistical difficulties in active mining environments. However, simulation offers a valuable solution by generating large and diverse datasets without incurring the risks and costs associated with real-world data collection.

Simulated environments provide researchers with precise control over data variability and complexity, facilitating the training and evaluation of algorithms specifically tailored for the detection of earth-moving equipment in mining contexts. Furthermore, simulations offer the opportunity for automatic annotation, significantly reducing the manual labor and costs involved in accurately labeling real-world data.

\begin{figure}
\includegraphics[width=\textwidth]{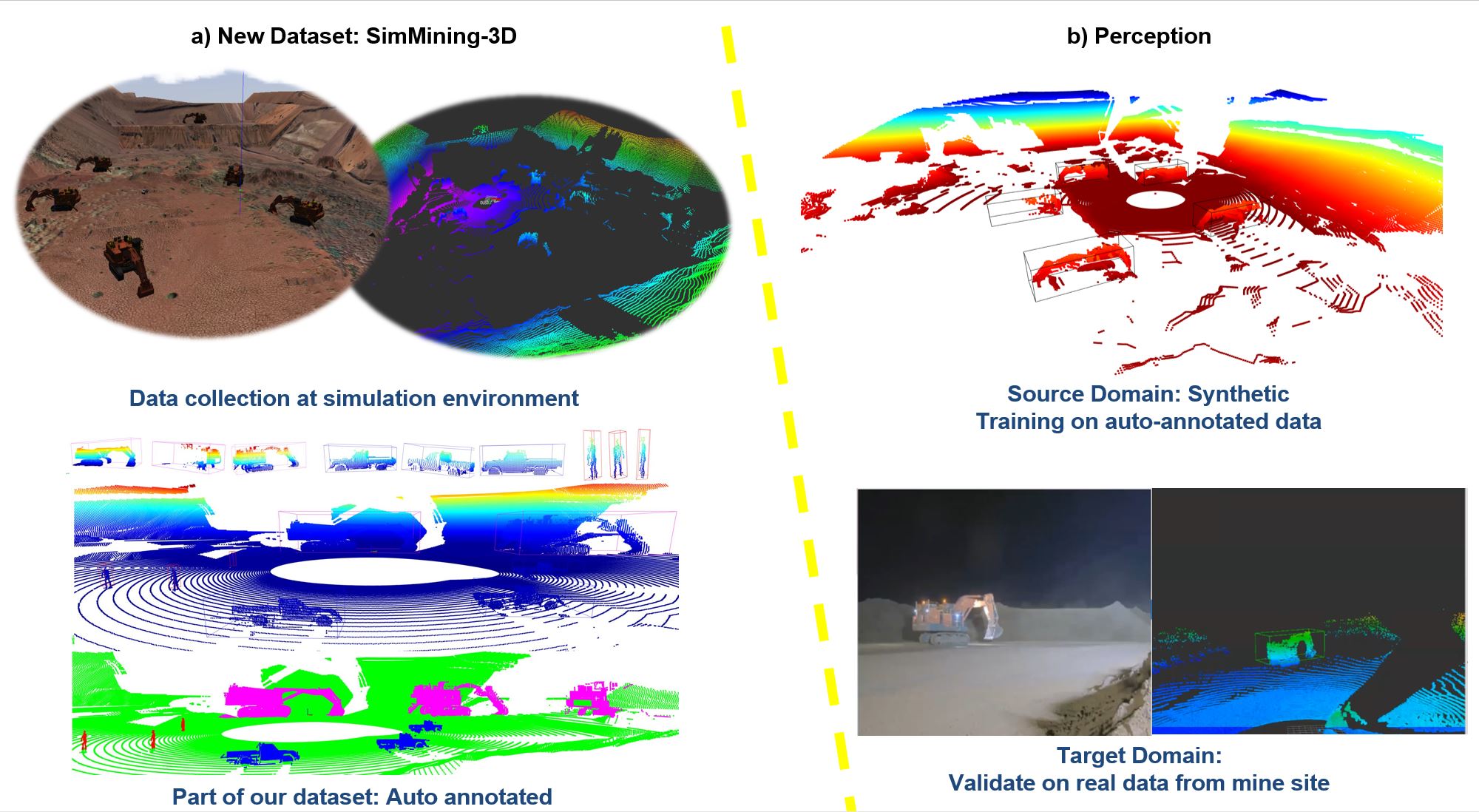}
\caption{Proposed workflow illustrates the scope of this study} \label{fig1}
\end{figure}
In this work, we propose a novel approach that addresses the domain shift between synthetic and real data in 3D object detection for complex mining environments. Our methodology involves training a model on a synthetic dataset generated from representative simulated mine environments. This approach effectively overcomes the challenges inherent to object detection in complex mining scenarios, accounting for variations in sensor height and other critical factors.

One of our significant contributions is the development of a comprehensive synthetic dataset explicitly designed for 3D object detection in mining environments. This dataset includes objects and sensors placed at different heights within pit benches, accurately capturing the complexities of real mining operations. Furthermore, we introduce novel evaluation metrics that consider sensor-to-object height and point cloud density variations for more accurate model performance assessment in mining scenarios.

Moreover, through extensive experiments, we successfully predict the presence of objects in real data captured from actual mining environments and evaluate the accuracy of our approach. The results demonstrate the effectiveness and practicality of our 3D object detection model in real-world mining scenarios.

Additionally, we have incorporated an automatic annotation pipeline leveraging the ROS interface. This pipeline includes new algorithmic solution to automate the annotation process, reducing manual labor and accelerating the dataset creation for 3D object detection in complex mining environments.

To assess the impact of altitude and height variation, we conduct an extensive ablation study. This study showcases the significance of an altitude shift augmentations in improving the overall accuracy and reliability of 3D object detection models specifically tailored for complex mining environments.

Our key contributions are as follows:

1) The development of a comprehensive synthetic dataset capturing the complexities of mining environments, including objects and sensors placed at different heights within pit benches.

2) Introduction of novel evaluation metrics that consider sensor-to-object height and point cloud density variations for more accurate model performance assessment in mining scenarios.

3) Successful prediction of objects in real data captured from actual mining environments, demonstrating the effectiveness of our approach.

4) The implementation of an automatic annotation pipeline using the ROS interface, significantly reducing manual labor and expediting the dataset creation process.

5) An extensive ablation study showcasing the importance of altitude and height variation augmentations in enhancing the accuracy and reliability of 3D object detection models for mining environments.

To support our research and foster collaboration, we have made our comprehensive synthetic dataset publicly accessible. Researchers can utilize this dataset as a benchmark for supervised learning, enabling the evaluation and advancement of object detection techniques in mining environments. Additionally, we provide a video summarizing our experimental trials, and the dataset is available at \cite{ref_article10}, ensuring accessibility and encouraging further exploration in this field.

By effectively bridging the gap between synthetic and real data, our work demonstrates the potential of synthetic data and simulation-based methodologies in overcoming domain shift challenges. We envision the development of robust and reliable object detection systems that can be practically deployed in mining and related domains, enhancing safety and operational efficiency.

\section{Related study }
Automatic annotation and data generation have been the focus of extensive research in recent years\cite{ref_article11}\cite{ref_article12}\cite{ref_article13}\cite{ref_article14}, with several state-of-the-art methods available for generating labeled data. One of the most popular techniques for data generation is through simulation environmentss. Simulations are widely used to collect data for machine learning applications, especially for perception systems. 

Gazebo \cite{ref_article15} and CARLA \cite{ref_article16} are two popular simulation environments used for autonomous vehicle research. These environments offer realistic virtual environments with a large number of sensors, including cameras and LiDARs, for data collection. Publicly available datasets from simulation environments have become a crucial source for training machine learning algorithms. Several datasets have been made publicly available for research purposes, including the CARLA \cite{ref_article16} , LGSVL\cite{ref_article17} , and Udacity  datasets\cite{ref_article18} , which include labeled data for autonomous vehicle perception systems. These datasets cover a wide range of scenarios, including urban and highway driving, and offer various sensor modalities, such as LiDAR, camera, and radar. In addition, these datasets offer accurate ground-truth labels for different perception tasks such as object detection, semantic segmentation, and lane detection. 

In \cite{ref_article19}\cite{ref_article20}, synthetic multimodal 3D raw data and automated semantic labeled data have been generated from Gazebo simulations of a ground vehicle operating in diverse natural environments and off-road terrains. The aim is to expedite software development and improve the generalization of models to new scenes.

However, the proposed approach for automatic 3D annotation aims to tackle the challenges of generating labeled data to increase automation in earth moving operations at mine sites. By leveraging the advantages of simulation environments, a large volume of labeled data can be generated. This approach has the potential to significantly enhance the performance of machine learning models for various applications in dynamic and degraded environments, such as mining.

\section{New Dataset: SimMining3D}
\subsection{Data Collection at simulated environment}
The data collection process took place within a representative simulation environment based on the Yandicoogina mine site, as described in \cite{ref_article21}. This simulated environment accurately replicates the real-world conditions and characteristics of the mine site, that contained within a rectangular area, spanning 583 meters in longitude and 379 meters in latitude. Additionally, the environment has an elevation of 63.5 meters, ensuring a comprehensive representation of the mine site's terrain and topography. The CAD models of earthmoving equipment and sensors were imported into Gazebo (Fig2).
\begin{figure}
\includegraphics[width=\textwidth]{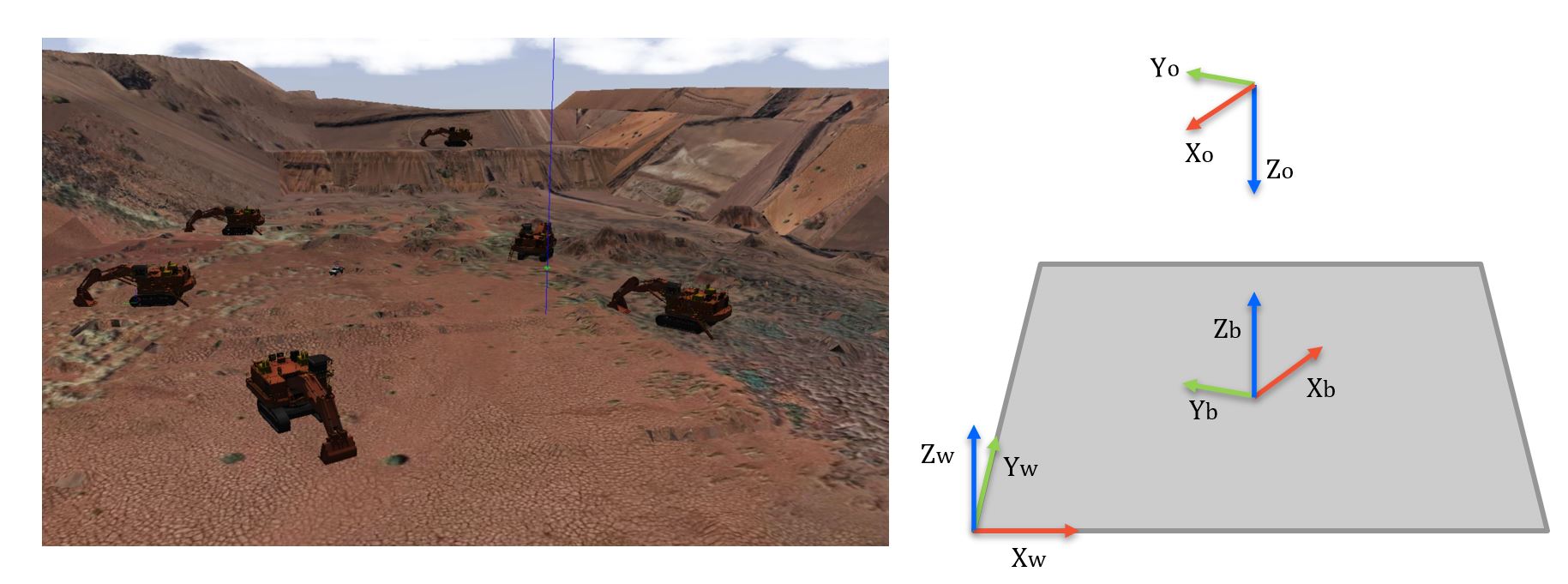}
\caption{Left: Representative Yandicoogina mine simulation environment in Gazebo, Right:Coordinate system used: x, y, z denote the coordinate system, with subscripts o, b, and w referring to Ouster sensor, mobile study trailer base footprint, and world coordinate, respectively.} \label{fig2}
\end{figure}
The point clouds in this dataset were acquired with a simulated MST system [reference ours]. The system consists of a simulated 128-line OS2 LiDAR sensor and a RGB camera. LiDAR sensor can capture point clouds at up to 700,000 points per second at a vertical field of view of 22.5°, with an accuracy of ±2.5 – 8 cm. 

Six excavators were placed at random locations in the simulation environment, and their movements and rotations were automatically controlled and repeated for each sensor position. The data collection of an object in the sensor coordinate frame is demonstrated in Figure 2.

The data collection process involved capturing information from distinct sensor heights in different scenarios. In one scenario, both the sensor and excavators were placed within the bench. In the second scenario, the sensor was positioned inside the pit, observing excavators within the pits and benches taller than 10 meters. In the third scenario, the sensor was placed on the benches, observing excavators within the pit. Excavators were continuously rotated at 0.2 radians per second and moved from their initial location by approximately 75m with a speed of 0.5m/s. In total, 818, 617, and 690 frames were captured for each scenario, respectively, resulting in a total of 2125 frames. This approach allowed us to gather diverse data across different sensor heights and excavator locations within the complex mining environment.

\subsection{Automatic Annotation}
Once the data is collected from the simulation environment, the next step in the automatic annotation pipeline is to determine the position, orientation, and dimension of the objects in the simulation environment. As detailed in the Algorithm 1 this information was obtained by using ROS tools and packages such as "tf" or "odometry" that provide the pose information of objects in the environment. Unlike other autono-mous vehicle or robots with sensors, the proposed data collection platform in this study will remain stable during data collection at multiple locations. Hence the base-footprint coordinate frame of the MST will not change constantly.
Then ROS package was used to publish the object pose information via object recognition messages. Using the published object pose information, the next step is to generate 3D bounding boxes around the objects. This was done by inferring the dimension of the objects and creating a box that encompasses the entire object from the ros messages as described in Algorithm 1. 
\begin{algorithm}
\caption{Automatic Annotation}
\begin{algorithmic}[1]
\State \textbf{Input:} Objects and sensor pose information, LiDAR scans, objects' dimensions (dx, dy, dz), Number of object classes=3
\State \textbf{Output:} kitty format.txt, Semantic Labels .csv, gt\_database
\ForAll{$f_i$ in frames}
  \State bbox = []
  \State pcd $\leftarrow$ read\_points($f_i$, field\_names=['x', 'y', 'z', 'rgb'])
  \ForAll{$o_i$ in objects}
    \State label $\leftarrow$ name($o_i$)
    \State center $\leftarrow$ center($o_i$, field\_names=['x', 'y', 'z'])
    \State size $\leftarrow$ Size($o_i$, field\_names=['dx', 'dy', 'dz'])
    \State rotation $\leftarrow$ RotationMatrix($o_i$, field\_names=['w', 'qx', 'qy', 'qz'])
    \State center[z] $\leftarrow$ center[z] + size[dz]/2
    \State roll\_X, pitch\_Y, yaw\_Z $\leftarrow$ quaternion\_to\_euler\_angle($o_i$, field\_names=['w', 'qx', 'qy', 'qz'])
    \State Write to text file ('x', 'y', 'z', 'dx', 'dy', 'dz', yaw\_Z, label)
  
    \State Bbox $\leftarrow$ OrientedBoundingBox(center, rot, size)
    \State Crop\_3d $\leftarrow$ crop(pcd, Bbox)
    \State Save\_point\_cloud(gt\_database/Crop\_3d\_$f_i$\_$o_i$, Crop\_3d)
    
    \ForAll{$c_i$ in [$c_1$, $c_2$, $c_3$]}
      \If{$o_i$ = $c_i$}
        \State color $\leftarrow$ [$r_i$, $g_i$, $b_i$]
        \State write to csv (frame\_$f_i$.csv, Crop\_3d (x, y, z, color, $c_i$))
      \EndIf
    \EndFor
  \EndFor
\EndFor
\end{algorithmic}
\end{algorithm}
The generated bounding box information were used to crop the point clouds of interested objects from the environment and stored in separates folders corresponding to each different objects as annotations in a format suitable for 3d object detection. Furthermore, as described in the original custom object format in the OpenPCDet library, the bounding box information in LiDAR coordinates, including the center of the box (x, y, and z), the dimensions of the box (dx, dy, and dz), yaw value, and the object class name, was saved for all objects in each scene. In addition, we introduced a new column in the above text files to indicate the objects' difficulties, labeled as 0 for easy, 1 for moderate, and 2 for hard difficulties. The corresponding changes were made in the OpenPCDet files to read and utilize this additional information from the text files. Similarly, the semantic values correspond to each 3D LiDAR points (coordinates) were saved in the csv files (Fig 3).

The annotation pipeline generated a large dataset of annotated objects in the simulation environment. To avoid close similarities between frames, we strategically selected total of 933 frames from continuous recordings, ensuring diversity in the dataset with corresponding labels provided with each difficulty level. The dataset includes ground truth information for 5353 excavators observed from multiple perspectives, ensuring comprehensive coverage and robust analysis. 

In addition to the 3D bounding box annotations used for object detection evaluation in this paper, complementary semantic point-wise labels were provided for 933 frames.

This dataset was used to train state of the art machine learning model for 3D object detection and semantic segmentation tasks. The dataset is available at \cite{ref_article10} 

\begin{figure}
\includegraphics[width=\textwidth]{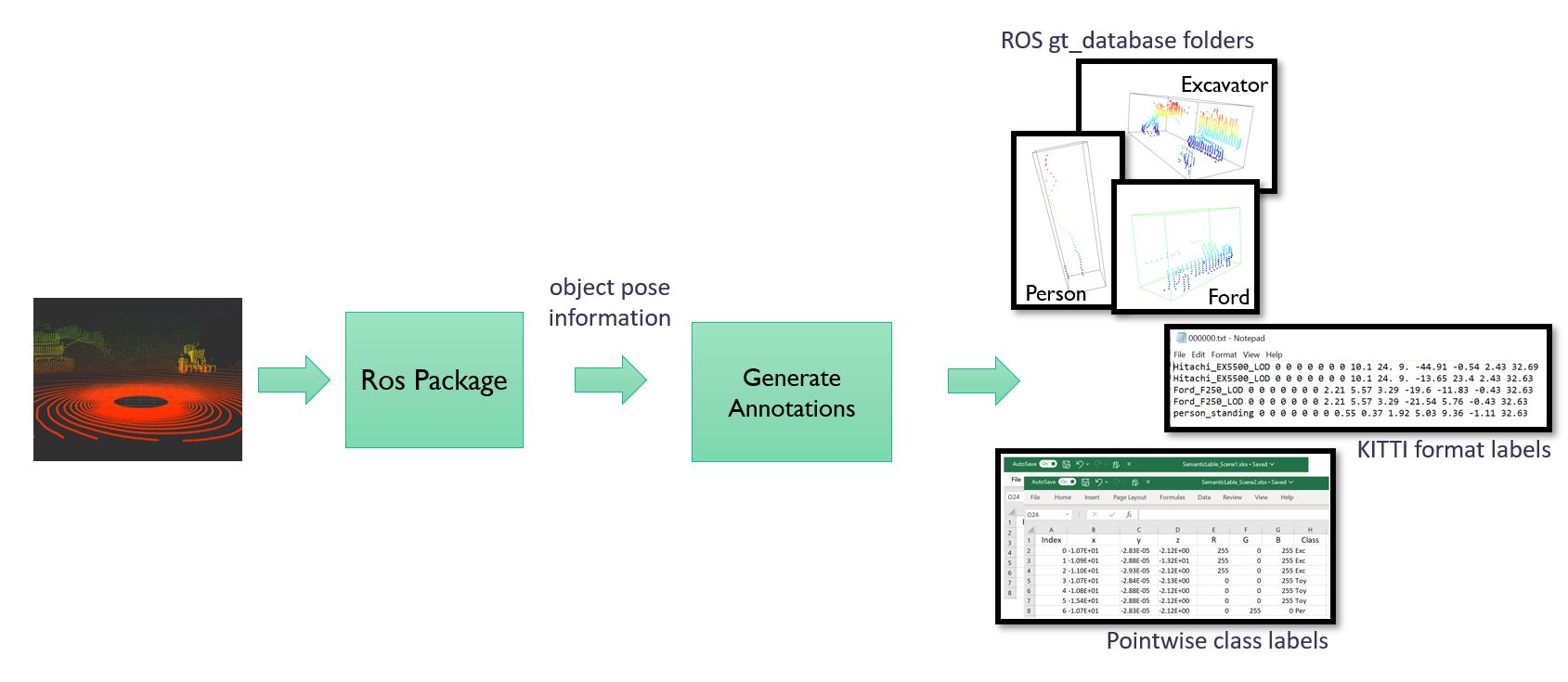}
\caption{Automatic annotation pipeline} \label{fig3}
\end{figure}

\section{Perception: Baseline Experiment}
\subsection{Experimental Setup}
Baseline results are provided using OpenPCDet framework \cite{ref_article22}, Poinpillar 3D object detection model trained on only the source domain of simulation environment, ideally representing the worst-case and best-case respectively for performance on the target domain-real data. PointPillars is an end-to-end model that uses a set of pillar-based representa-tions for 3D object detection in point clouds. It processes each pillar individually and applies a set of convolutions and max-pooling operations to generate features for object detection\cite{ref_article23}. 

This study focuses on the exacavator static model. Unlike in other 3d object detection models studied in other literature the excavators are dimensionally huge and the small change in the 3d box orientation can result noticeably different in the prediction. Given the large discrepancy in the size of the ground truth anchor boxes and the points from the objects, we filter out the ground truths which consists of less than 100 points from the objects.

As our datasets contain only one object class that needs to be detected, we adapt the output layers of the algorithms to predict a single class only. The anchor size is set to the ground truth dimensions of the Hitachi excavator, with l = 8.65m, w = 23.9m, and h = 10.02m for length l, width w, and height h, respectively. We empirically test different network parameters, such as the voxel size for PointPillars, the number of filters in network. For all other parameters, we choose the voxel size of 2.19, 2.19, 14 and the maximum point per voxels of 32. For data augmentation, we use random horizontal flip, scale augmentations. Although the LiDAR sensors capture reflections at over 200 m distance, we limit the detection range of our network to a horizontal range of 175.2m in the dimensions x and y and -12 and 4m in z direction in order to capture the large variation of mine pit walls and the tall objects. We remove the intensity channel from network and only use x, y, and z as input features. The network was trained on 1235 object samples chosen from all senarios with batch size of 2 and learning rate for 120 epochs and validated on 509 object samples using modified-Kitty format validation as discussed at section Evaluation metric.

\subsubsection{Data Augmentation} 
In this study, we propose the 'altitude shift' augmentation technique tailored for point cloud data. Unlike the previous Random World Translation (RWT) method, our approach utilizes a uniform distribution to generate random values within a specified range. By simulating altitude variations, the altitude shift augmentation enhances the adaptability and generalization capabilities of 3D object detection models.

Unlike RWT technique, which introduces noise based on standard deviations for each axis, our Altitude Shift augmentation guarantees an equal probability for all values within the specified range. This approach allows the models to adapt to various object heights commonly found in real-world scenarios. Additionally, the Altitude Shift is implemented on-the-fly during each run, providing the flexibility to either apply a fixed shift (Constant\_Altitude\_Shift) or randomly sample values (Random\_Altitude\_Shift) within the range.

Implemented along the z-axis, the altitude shift function modifies the vertical position of points and their corresponding ground truth bounding boxes. It takes three input parameters: ground truth boxes (gt\_box), points, and an offset range determining the permissible shift values.
\begin{algorithm}
\caption{Altitude Shift}
\begin{algorithmic}[1]
\State  \textbf{Input:} LiDAR point cloud $L \in \mathbb{R}^{N \times 3}$ with $N$ points, ground truth box $gt\_box$, Offset range $[\mathrm{min}, \mathrm{max}]$
\State  \textbf{Output:}Shifted LiDAR point cloud $L$

\For{$l \in L$}
  \State offset $\gets$ RandomUniform($\mathrm{offset\_range}[0]$, $\mathrm{offset\_range}[1]$)
  \State $l.z \gets l.z + \mathrm{offset}$
\State $gt\_box.z \gets gt\_box.z  + \mathrm{offset}$
\EndFor

\end{algorithmic}
\end{algorithm}

We further evaluate other augmentation techniques based on OpenPCDet \cite{ref_article22} to compare the impact of altitude variation on the overall accuracy in complex object detection environments. These environments are characterized by varying heights, presenting unique challenges for accurate object detection. By conducting a comprehensive analysis, we aim to understand how different augmentation strategies, including altitude variation, specifically address the complexities associated with height variation.

\subsubsection{Evaluation Metric} 
We assess excavator detection difficulty using specific criteria based on height variation between the sensor and object, as well as point cloud density. The difficulty levels are as follows:

Level 0 - Easy: Height variation < 10m, point cloud density > 750 points.

Level 1 - Moderate: Height variation < 10 units, point cloud density 100 to 750 points, or height variation > 10m, point cloud density > 750 points.

Level 2 - Hard: Height variation > 10m, point cloud density 100 to 750 points.

By categorizing excavator instances into these levels, we evaluate detection algorithm performance in complex mining environments. Our modifications in OpenPCDet enable accurate evaluation aligned with mining-specific challenges.

\subsection{Results and Discussion} 

\subsubsection{Evaluations on Simulated data}
The evaluation was conducted on a synthetic dataset captured in a complex mining environment using a simulation environment. The 3D object detection performance of Pointpillar for the Hitachi excavator is presented in Table \ref{tab1}, reporting Mean Average Precision (mAP) results for 3D and Bird's Eye View (BEV) detection at IoU 0.7. The evaluation includes 40 recall positions, providing comprehensive assessment across difficulty levels: Easy, Moderate, and Hard proposed in this paper. The results demonstrate the impact of different augmentation techniques compared to no augmentation on object detection performance in complex environments. 
\begin{table}[htbp]
\centering
\caption{Object Detection Performance Evaluation with Augmentation (BEV and 3D at 0.7 IoU)}
\label{tab1}
\begin{tabular}{|l|c|c|c|c|c|c|}
\hline
\textbf{Method} & \multicolumn{3}{c|}{\textbf{BEV at 0.7 IoU}} & \multicolumn{3}{c|}{\textbf{3D at 0.7 IoU}} \\
\cline{2-7}
 & \textbf{Easy} & \textbf{Mod.} & \textbf{Hard} & \textbf{Easy} & \textbf{Mod.} & \textbf{Hard} \\
\hline
None & 94.96 & 90.44 & 61.05 & 91.57 & 87.38 & 58.23 \\
RAS ([-2,2]) & 99.00 & 94.16 & 64.35 & 94.31 & 90.10 & 60.86 \\
RAS([-0.5,0.5]) & 95.42 & 91.00 & 61.40 & 94.45 & 87.93 & 60.91 \\
CAS (0.5) & 93.23 & 86.92 & 60.25 & 56.38 & 35.19 & 37.73 \\
RWT\_z\_only (0.5) & 95.47 & 91.00 & 61.47 & 88.64 & 84.65 & 57.93 \\
Standard & 98.11 & 91.20 & 61.59 & 97.85 & 91.00 & 61.43 \\
Standard + RAS & 98.47 & 91.33 & 61.65 & 94.52 & 90.12 & 60.82 \\
Standard + CAS & 97.94 & 90.97 & 61.45 & 85.66 & 79.82 & 53.88 \\
Standard + RWT\_z\_only & 97.79 & 90.93 & 61.39 & 93.88 & 87.37 & 60.56 \\
\hline
\end{tabular}
\end{table}
The Random\_Altitude\_Shift (RAS) augmentation is highly effective for improving object detection accuracy. It consistently outperforms the baseline across difficulty levels and evaluation metrics, providing significant improvements in both BEV and 3D detection.
Comparing RAS with other techniques like RWT, CAS, and standard augmentation, RAS performs better, especially in the Hard level. It captures altitude variations in the dataset, resulting in more accurate object detection. CAS and standard augmentation show improvements but are not as consistent as RAS. RWT, focusing on z-axis offset, lags behind RAS in accuracy. 
Considering altitude-specific variations is crucial for improved detection, as shown by RAS outperforming RWT. The controlled altitude shifts introduced by RAS allow better adaptation to real-world height variations. RAS has practical implications for robust object detection in complex environments, enhancing accuracy and reliability in critical applications like autonomous driving and robotics.
\subsubsection{Validation on real data}
Predictions on both simulated and real data were presented in Figure 4. Figure 4(b) and (c) showcase the video and point cloud captured at the real minesite, respectively. Figure 4(c) demonstrates the successful transfer of the trained model on synthetic data to a real-world scenario. Considering the disparity in lidar positions between simulation and reality in terms of height, the input point cloud was transformed to accommodate the sensor's height variation during model prediction. Videos depicting the predictions from simulation to simulation and simulation to real for various scenarios can be found at \cite{ref_article10}.
\begin{figure}
\includegraphics[width=\textwidth]{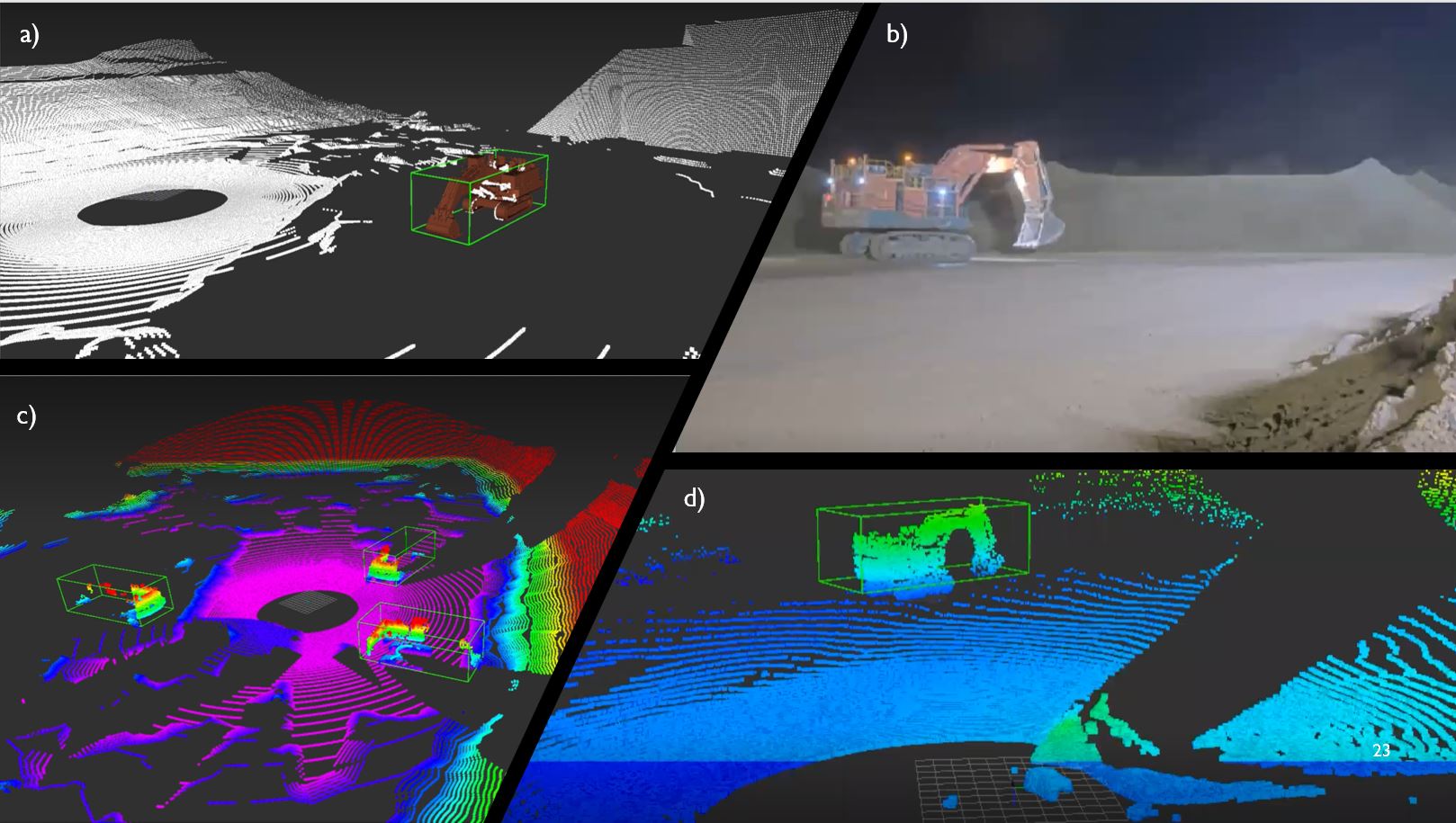}
\caption{Model predictions on both simulated data (a),(c) and real data (d)} \label{fig4}
\end{figure}
\section{Conclusion}  
In conclusion, our study emphasizes the significant contributions of synthetic data generation, automatic annotation, altitude shift augmentation, and sim-to-real transformation in enhancing object detection models for mining environments.

Furthermore, excavators' complex shapes during operations should be accommodated in future studies. Addressing the impact of random values on augmentations through systematic work will provide valuable insights.

The integration of these approaches improves the accuracy and reliability of object detection models for mining applications. These advancements have the potential to enhance safety, efficiency, and productivity in mining operations, addressing challenges related to domain gap and limited real data availability.

\subsubsection{Acknowledgements} This work has been supported by the Australian Centre for Robotics and the Rio Tinto Centre for Mine Automation, the University of Sydney.
\\
\\
\\
\\
\\
\\
%
%
%

\begin{thebibliography}{9}
\bibitem{ref_article10}
SimMining-3D:\url{https://github.com/MehalaBala/SimMining\_3D}
\bibitem{ref_article11}
Nikolenko, S. Synthetic Simulated Environments. In Synthetic Data for Deep Learning; Springer Optimization and Its Applications; Springer: Cham, Switzerland, 2021; Volume 174, Chapter 7; pp. 195–215.
\bibitem{ref_article12}
Yue, X.; Wu, B.; Seshia, S.A.; Keutzer, K.; Sangiovanni-Vincentelli, A.L. A LiDAR Point Cloud Generator: From a Virtual World to Autonomous Driving. In Proceedings of the ACM International Conference on Multimedia Retrieval, Yokohama, Japan, 11–14 June 2018; pp. 458–464.
\bibitem{ref_article13}
Smith, Amos et al. “A Deep Learning Framework for Semantic Segmentation of Underwater Environments.” OCEANS 2022, Hampton Roads (2022): 1-7.
\bibitem{ref_article14}
R. P. Saputra, N. Rakicevic and P. Kormushev, "Sim-to-Real Learning for Casualty Detection from Ground Projected Point Cloud Data," 2019 IEEE/RSJ International Conference on Intelligent Robots and Systems (IROS), Macau, China, 2019, pp. 3918-3925, doi: 10.1109/IROS40897.2019.8967642.
\bibitem{ref_article15}
K. Koenig and A. Howard, "Design and use paradigms for Gazebo an open-source multi-robot simulator", Proc. IEEE-RSJ International Conference on Intelligent Robots and Systems, pp. 2149-2154, 2004.
\bibitem{ref_article16}
D. Dworak, F. Ciepiela, J. Derbisz, I. Izzat, M. Komorkiewicz and M. Wójcik, "Performance of LiDAR object detection deep learning architectures based on artificially generated point cloud data from CARLA simulator," 2019 24th International Conference on Methods and Models in Automation and Robotics (MMAR), Miedzyzdroje, Poland, 2019, pp. 600-605, doi: 10.1109/MMAR.2019.8864642.
\bibitem{ref_article17}
G. Rong et al., "LGSVL Simulator: A High Fidelity Simulator for Autonomous Driving," 2020 IEEE 23rd International Conference on Intelligent Transportation Systems (ITSC), Rhodes, Greece, 2020, pp. 1-6, doi: 10.1109/ITSC45102.2020.9294422.
\bibitem{ref_article18}
“Udacity Dataset,” https://github.com/udacity/self-driving-car/tree/master/datasets, 2018.
\bibitem{ref_article19}
Sánchez, M.; Morales, J.; Martínez, J.L.; Fernández-Lozano, J.J.; García-Cerezo, A. Automatically Annotated Dataset of a Ground Mobile Robot in Natural Environments via Gazebo Simulations. Sensors 2022, 22, 5599. https://doi.org/10.3390/s22155599
\bibitem{ref_article20}
A. Tallavajhula, Ç. Meriçli and A. Kelly, "Off-Road Lidar Simulation with Data-Driven Terrain Primitives," 2018 IEEE International Conference on Robotics and Automation (ICRA), Brisbane, QLD, Australia, 2018, pp. 7470-7477, doi: 10.1109/ICRA.2018.8461198.
\bibitem{ref_article21}
Balamurali, M. and Hill, A. J. and Martinez, J. and Khushaba, R. and Liu, L. and Kamyabpour, N. and Mihankhah, E. A Framework to Address the Challenges of Surface Mining through Appropriate Sensing and Perception, 17th International Conference on Control, Automation, Robotics and Vision (ICARCV),2022, 261-267, 10.1109/ICARCV57592.2022.10004309
\bibitem{ref_article22}
OpenPCDet Development Team. Openpcdet: An opensource toolbox for 3d object detection from point clouds.https://github.com/open-mmlab/OpenPCDet,2020. 5
\bibitem{ref_article23}
A. H. Lang, S. Vora, H. Caesar, L. Zhou, J. Yang, and O. Beijbom.Pointpillars: Fast encoders for object detection frompoint clouds. CoRR, abs/1812.05784, 2018. 2, 4


\end{thebibliography}
%
%
%
%

\end{document}